\documentclass[10pt,twocolumn,letterpaper]{article}
\usepackage{cvpr}
\usepackage{times}
\usepackage{epsfig}
\usepackage{graphicx}
\usepackage{amsmath}
\usepackage{amssymb}
\usepackage{subfig}
\usepackage{authblk}
\usepackage{cite}
\usepackage{comment}
\usepackage{adjustbox}
\usepackage{textcomp}
\usepackage{multirow}
\usepackage[table]{xcolor}
\usepackage{xcolor,colortbl}
\usepackage{bbm}
\usepackage{makecell}

\usepackage[pagebackref=true,breaklinks=true,letterpaper=true,colorlinks,bookmarks=false]{hyperref}

\cvprfinalcopy % *** Uncomment this line for the final submission

\def\cvprPaperID{221} % *** Enter the CVPR Paper ID here

\begin{document}

%%%%%%%%% TITLE

%\title{Deep Encoding Network for Texture Recognition}
% Maybe new title
\title{Deep TEN: Texture Encoding Network}

\author[ ]{
Hang Zhang
}
\author[ ]{
Jia Xue
}
\author[ ]{
Kristin Dana
}
\affil[ ]{Department of Electrical and Computer Engineering, Rutgers University, Piscataway, NJ 08854}
\affil[ ]{ {\tt\small \{zhang.hang,jia.xue\}@rutgers.edu, kdana@ece.rutgers.edu }}
\renewcommand\Authsep{  } 
\renewcommand\Authands{  }

\maketitle
%\thispagestyle{empty}

%%%%%%%%% ABSTRACT
%-------------------------------------------------------------------------
\begin{abstract}
We propose a Deep Texture Encoding Network (Deep-TEN) with a novel Encoding Layer integrated on top of convolutional layers, which ports the entire dictionary learning and encoding pipeline into a single model.  
Current methods  build from distinct components, using standard encoders with separate off-the-shelf features such as SIFT descriptors or pre-trained CNN features for material recognition.
Our new approach provides an end-to-end learning framework,  where the inherent visual vocabularies are learned directly from the loss function.  The  features,  dictionaries and the encoding representation for the classifier are all learned simultaneously. 
The representation is orderless and therefore is particularly useful for material and texture recognition. The Encoding Layer generalizes robust residual encoders 
such as VLAD and Fisher Vectors, and has the property of discarding domain specific information which makes the learned convolutional features easier to transfer.
Additionally, joint training using multiple datasets of varied sizes and class labels is supported resulting in increased recognition performance. 
The experimental results show superior performance as compared to state-of-the-art methods using gold-standard databases such as MINC-2500, Flickr Material Database, KTH-TIPS-2b, and two recent databases 4D-Light-Field-Material and GTOS.  
The source code for the complete system are publicly
available\footnote{\url{http://ece.rutgers.edu/vision}}.

\end{abstract}

%%%%%%%%% BODY TEXT
%-------------------------------------------------------------------------
\section{Introduction}

\begin{figure}[t]
\includegraphics[width=\linewidth]{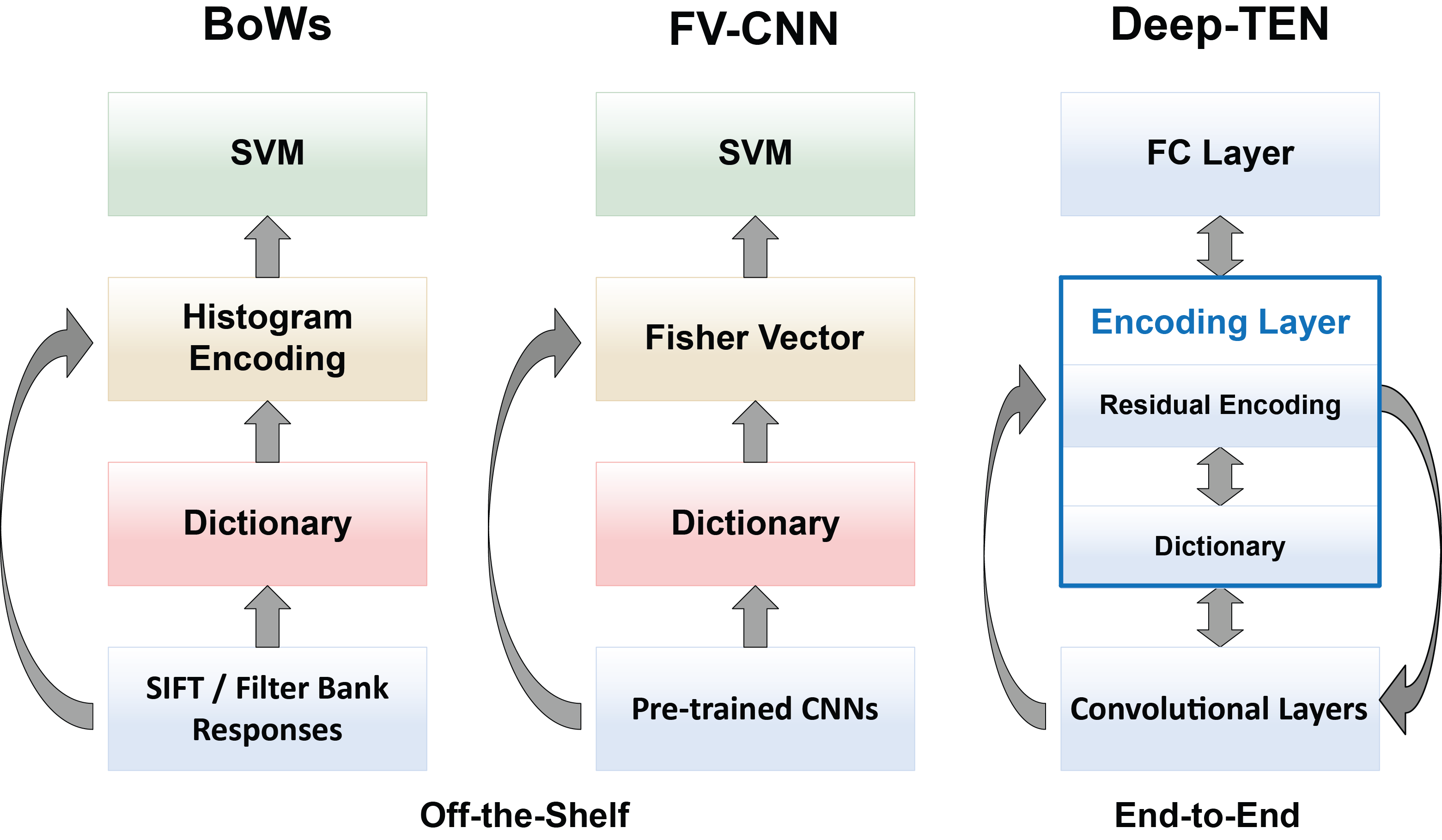}
\caption{ A comparison of classic approaches and the proposed Deep Texture Encoding Network. Traditional methods such as bag-of-words BoW (left) have a structural similarity to more recent FV-CNN methods (center). Each component is optimized in separate steps as illustrated with different colors. In our approach (right) the entire pipeline is learned in an integrated manner, tuning each component for the task at hand (end-to-end texture/material/pattern recognition).}
\label{fig:overview}
\end{figure}

% motivations and applications are needed
%Material recognition is essential for a machine to understand and interact with the world, for example a robot can easily grab an object if it knows what it is made of and a vehicle can automatically switch driving modes based on the ground terrains (asphalt, sand, grass or ice). 
With the rapid growth of deep learning,
 convolutional neural networks (CNNs) has become the de facto standard in many
object recognition algorithms.
The goals of material and texture recognition algorithms, while similar to object recognition, have the distinct challenge of capturing an orderless measure encompassing some spatial repetition. For example, distributions or histograms of features provide an orderless encoding for recognition. 
In classic computer vision approaches for material/texture recognition,
hand-engineered features are extracted using interest point detectors such as  SIFT~\cite{lowe2004distinctive} or filter bank responses~\cite{Cula01b,Cula01a,leung2001representing,varma2002classifying}. 
A dictionary is typically learned offline and then the feature distributions are encoded by Bag-of-Words (BoWs)~\cite{joachims1998text,csurka2004visual,sivic2005discovering,fei2005bayesian},
 In the final step, a classifier such as SVM is learned for classification. 
In recent work, hand-engineered features and filter banks are replaced by pre-trained CNNs and BoWs are replaced by the robust residual encoders such as VLAD~\cite{jegou2010aggregating} and its probabilistic version Fisher Vector (FV)~\cite{perronnin2010improving}. For example, Cimpoi {\it et al.}~\cite{cimpoi15} assembles different features (SIFT, CNNs) with different encoders (VLAD, FV) and have achieved state-of-the-art results. 
These existing approaches have the advantage of accepting arbitrary input image sizes and have no issue when transferring features across different domains since the low-level features are generic. 
However, these  methods (both classic and recent work) are comprised of stacking self-contained algorithmic components (feature extraction, dictionary learning, encoding, classifier training) as visualized in Figure~\ref{fig:overview} (left, center). Consequently, they have the disadvantage that the features and the encoders are fixed once built, so that feature learning (CNNs and dictionary) does not benefit from labeled data. We present a new approach (Figure~\ref{fig:overview}, right) where the entire pipeline is learned in an end-to-end manner. 

Deep learning ~\cite{krizhevsky2012imagenet} is well known as an end-to-end learning of hierarchical features, so what is the challenge in recognizing textures in an end-to-end way? 
The convolution layer of CNNs operates in a sliding window manner acting as a local feature extractor. The output featuremaps preserve a relative spatial arrangement of input images. 
The resulting globally ordered features are then concatenated and fed into the FC (fully connected) layer which acts as a classifier. 
This framework has achieved great success in image classification, object recognition, scene understanding and many other applications, but is typically not ideal for recognizing textures due to the need for an spatially invariant representation describing the feature distributions instead of concatenation.
Therefore, an orderless feature pooling layer is desirable for
%Since an orderless representation is desirable, we want a new 
end-to-end learning. % that integrates orderless encoding. 
The challenge is to make the loss function differentiable with respect to the  inputs and layer parameters. We derive a new back propagation equation series (see Appendix~\ref{sec:appA}). In this manner, encoding for an orderless representation can be integrated within the deep learning pipeline. 
%An added benefit is that input size can be arbitrary and the learned features are transferrable,  while existing CNNs have the limitation of requiring a fixed input size and the learned features do not transfer easily due to domain-specific information. 

% main contribution of this paper
As the {\bf first contribution} of this paper, we introduce a novel {\it learnable residual encoding layer} which we refer to as the {\it Encoding Layer}, that ports the entire dictionary learning and residual encoding pipeline into a single layer for CNN. 
%that integrates the powerful dictionary encoding framework with the richness of deep CNN features.
%The Encoding Layer generalizes dictionary learning and residual encoding representations. 
The Encoding Layer has three main properties. (1) The Encoding Layer generalizes robust residual encoders such as VLAD and Fisher Vector. This representation is orderless and describes the feature distribution, which is suitable for material and texture recognition. 
(2) The Encoding Layer acts as a pooling layer integrated on top of convolutional layers, accepting arbitrary input sizes and providing output as a fixed-length representation.
By allowing arbitrary size images, the Encoding Layer makes the deep learning framework more flexible and our experiments show that recognition performance is often improved with multi-size training. 
In addition,  (3) the Encoding Layer learns an inherent dictionary and the encoding representation which is likely to carry domain-specific information and therefore is suitable for transferring pre-trained features.
In this work, we transfer  CNNs from object categorization (ImageNet\cite{imagenet}) to material recognition.
Since the network is trained end-to-end as a regression progress, the convolutional features learned together with Encoding Layer on top are easier to transfer (likely to be domain-independent). 

%the Encoding Layer has the property of discarding the influence of frequently appearing patterns that are typically  domain-specific. 
%Our experiments show that the Encoding Layer has superior performance for transferring pre-trained CNNs. 

The {\bf second contribution} of this paper is a new framework for end-to-end material recognition which we refer to as {\it Texture Encoding Network - Deep TEN},  where the feature extraction, dictionary learning and encoding representation are learned together in a single network as illustrated in Figure~\ref{fig:overview}.
%building a deep convolutional network with an Encoding Layer which we refer to as {\it Texture Encoding Network - Deep TEN} which learns the discriminative CNN features (convolutional layer) and the encoding representation in an end-to-end manner as shown in Figure~\ref{fig:overview} (right). 
Our approach has the benefit of gradient information passing to each component during back propagation, tuning each component for the task at hand. 
Deep-Ten outperforms existing modular methods and achieves the {\bf state-of-the-art} results on material/texture datasets such as MINC-2500 and KTH-TIPS-2b.  Additionally,  this Deep Encoding Network performs well in general recognition tasks beyond texture and material as demonstrated with results on MIT-Indoor and Caltech-101 datasets. 
We also explore how convolutional features learned with Encoding Layer can be  transferred through joint training on two different datasets. The experimental result shows that the recognition rate is significantly improved with this joint training.

\section{Learnable Residual Encoding Layer}
\label{sec:encoding}

\begin{figure}[t]
\begin{center}
\includegraphics[width=0.95\linewidth]{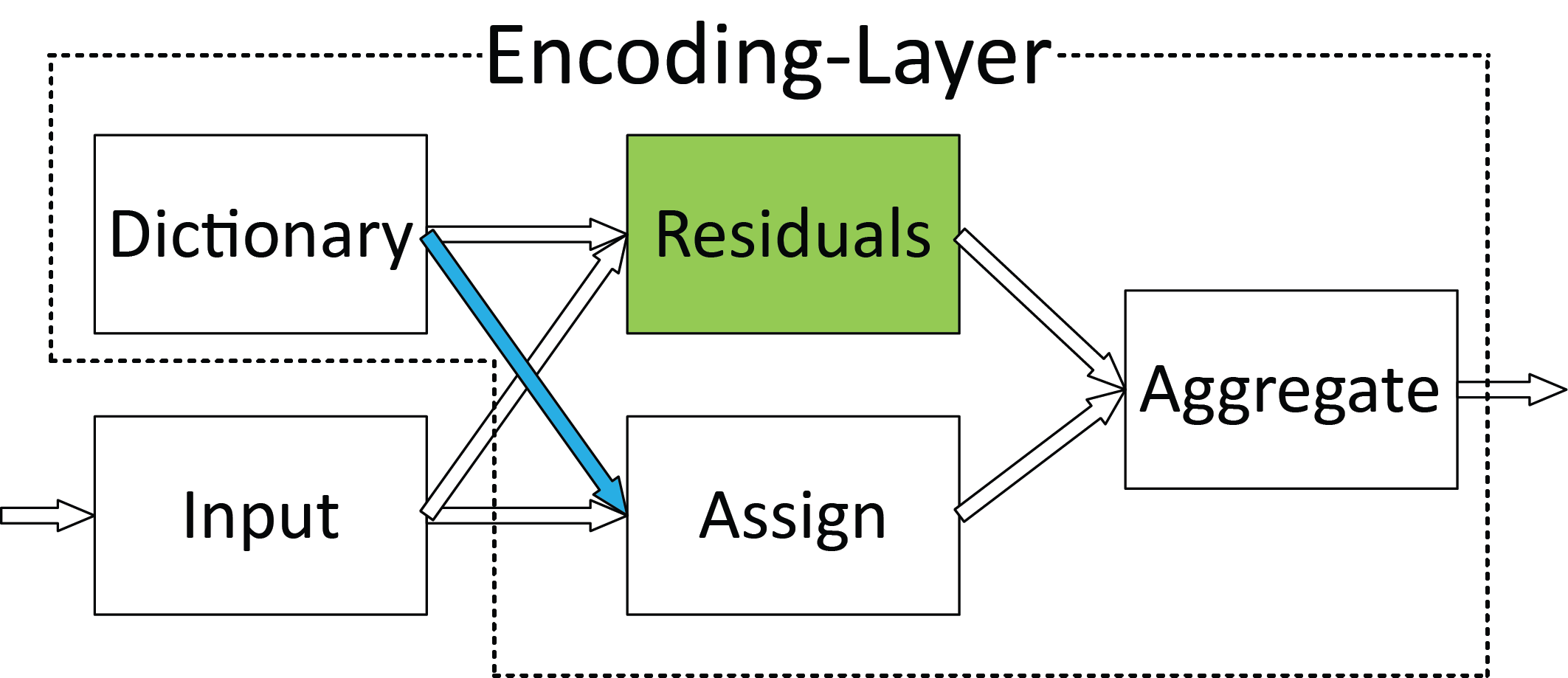}
\end{center}
\caption{ 
The {\it Encoding Layer} learns an inherent {\it Dictionary}.
The {\it Residuals} are calculated by pairwise difference between visual
descriptors of the input and the codewords of
the dictionary.  Weights are {\it assigned}  based on  pairwise
distance between descriptors and codewords. Finally, the residual vectors are {\it aggregated}  with 
the assigned weights. 
}
\label{fig:encoding}
\end{figure}

\definecolor{TableRed}{HTML}{800000}
\newcommand{\TextRed}[1]{\textcolor{TableRed}{#1}}
\begin{table*}[t]
\centering
\begin{adjustbox}{max width=\textwidth}
\begin{tabular}{|l|c|c|c|c|c|c|c|c|c|}
 \hline
   & Deep Features & Dictionary Learning & Residual Encoding & 
   Any-size & Fine-tuning & End-to-end Classification \\
 \hline
  BoWs & & \checkmark & & \checkmark & & \\
 \hline
  Fisher-SVM~\cite{sydorov2014deep} & & \checkmark & \checkmark & \checkmark & & \\
 \hline
 Encoder-CNN \small{(FV~\cite{cimpoi15} VLAD~\cite{gong2014multi}} & \checkmark & \checkmark & \checkmark & \checkmark &  & \\
 \hline
  CNN & \checkmark & & &  & \checkmark & \checkmark \\
 \hline
 B-CNN\cite{lin2015bilinear} & \checkmark & & & & \checkmark &  \\
 \hline
  SPP-Net\cite{he2014spatial} & \checkmark & & & \checkmark & \checkmark & \checkmark \\
 \hline
 Deep TEN \small{(\TextRed{ours})} & \checkmark & \checkmark & \checkmark & \checkmark & \checkmark & \checkmark\\
 \hline
\end{tabular}
\end{adjustbox}
\caption{Methods Overview. Compared to existing methods, Deep-Ten has several desirable properties: it integrates deep features with  dictionary learning and residual encoding and it allows any-size input, fine-tuning and provides end-to-end classification.  }
\label{tab:compare}
\end{table*}

% Hang, no need to give an outline of the section
%In this section, we first introduce a general formula forresidual encoding model, which generalizes the residual encoders. Then we proposethe Encoding Layer, which ports the whole dictionary learning and encoding framework into a single learnable layer for deep learning. Comparisons with existing methods and representations are provided.

\paragraph{\bf Residual Encoding Model} 
Given a set of $N$ visual descriptors $X=\{x_1,..x_N\}$ and a learned codebook $C=\{c_1, ... c_K\}$ containing $K$ codewords that are $D$-dimensional, each descriptor $x_i$ can be assigned with a weight $a_{ik}$ to each codeword $c_k$ and the corresponding residual vector is denoted by $r_{ik}=x_i-c_k$, where $i=1,...N$ and $k=1,...K$. 
Given the assignments and the residual vector, the residual encoding model applies an aggregation operation for every single codeword $c_k$:
\begin{equation}
e_k=\sum_{i=1}^N e_{ik}=\sum_{i=1}^N a_{ik}r_{ik} .
\label{eq:aggregation}
\end{equation}

The resulting encoder outputs a fixed length representation $E=\{e_1,...e_K\}$ (independent of the number of input descriptors $N$). 

\paragraph{Encoding Layer}
The traditional visual recognition approach can be partitioned into feature extraction, dictionary learning, feature pooling (encoding) and classifer learning as illustrated in Figure~\ref{fig:overview}. 
% If the reviewer is an author of these methods the following line is too harsh
%The multi-step methods are tedious and inelegant.
%Inspired by residual encoding, 
In our approach, we port the dictionary learning and
residual encoding
 into a single layer of CNNs, which we refer to as 
the {\it Encoding Layer}. The Encoding Layer simultaneously 
learns the encoding parameters along with with an inherent dictionary 
in a fully supervised manner.
The inherent dictionary is learned from the distribution of 
the descriptors by passing the gradient through assignment weights. 
During the training process, 
the updating of extracted convolutional
features can also benefit from the encoding representations.

Consider the assigning weights for assigning the descriptors to the codewords. Hard-assignment provides a single non-zero assigning weight for each descriptor $x_i$, which corresponds to the nearest codeword.
The $k$-th element of the assigning vector is given by $a_{ik}=\mathbbm{1} (\|r_{ik}\|^2=min\{\|r_{i1}\|^2,...\|r_{iK}\|^2\})$
where $\mathbbm{1}$ is the indicator function (outputs 0 or 1).
%where $\delta()$ is an indicator function and $j$ satisfies. 
Hard-assignment doesn't consider the codeword ambiguity and also
makes the model non-differentiable.
Soft-weight assignment addresses this issue by assigning a descriptor
to each codeword \cite{Gemert08}. The assigning weight is given by
\begin{equation}
a_{ik}=
%\frac{\exp(-\beta \|x_i-c_k\|^2)}{\sum_{j=1}^K\exp(-\beta \|x_i-c_j\|^2)} ,
\frac{\exp(-\beta \|r_{ik}\|^2)}{\sum_{j=1}^K\exp(-\beta \|r_{ij}\|^2)} ,
\end{equation}
where $\beta$ is the smoothing factor for the assignment.

Soft-assignment assumes that different clusters have equal scales.
Inspired by guassian mixture models (GMM), 
we further allow the smoothing factor $s_k$ 
for each cluster center $c_k$ to be learnable:
\begin{equation}
\label{eq:assigning}
a_{ik}=
\frac{\exp(-s_k \|r_{ik}\|^2)}{\sum_{j=1}^K\exp(-s_j \|r_{ij}\|^2)} ,
\end{equation}
which provides a finer modeling of the descriptor
distributions.
%Talk about joint learning the assignment and the codewords.
The Encoding Layer concatenates the aggregated
residual vectors with assigning weights 
(as in Equation~\ref{eq:aggregation}). 
As is typical in prior 
work~\cite{perronnin2010improving,arandjelovic2015netvlad}, 
the resulting vectors are normalized 
using the $L2$-norm.

\paragraph{\bf End-to-end Learning}
The Encoding Layer is a directed acyclic 
graph as shown in Figure~\ref{fig:encoding}, and all the components 
are differentiable {\it w.r.t} the input $X$ and the parameters
(codewords $C=\{c_1,...c_K\}$ and smoothing factors $s=\{s_1,...s_k\}$).
Therefore, the Encoding Layer can be trained
end-to-end by standard SGD (stochastic gradient descent) with backpropagation. 
%the gradient {\it w.r.t} the input features $X$ can be back-propagated, 
%and the codewords $C=\{c_1,...c_K\}$ and smoothing factors  $s=\{s_1,...s_k\}$ can be learned by standard Stochastic Gradient Descent (SGD).
We provide the 
details and relevant equation derivations in the Appendix~\ref{sec:appA}.

\subsection{Relation to Other Methods}

\paragraph{\bf Relation to Dictionary Learning}
Dictionary Learning is usually learned from the distribution 
of the descriptors in an unsupervised manner. 
K-means~\cite{lloyd1982least} learns the dictionary using 
hard-assignment grouping. 
Gaussian Mixture Model (GMM)~\cite{everitt1981finite} is 
a probabilistic version of K-means, which
allows a finer modeling of the feature distributions. 
Each cluster is modeled by a Gaussian component with its own mean,
variance and mixture weight.
The Encoding Layer makes the inherent dictionary differentiable {\it w.r.t}
the loss function and learns the dictionary in a supervised manner.
To see the relationship of the Encoding Layer to K-means, consider Figure~\ref{fig:encoding} with 
omission of the residual vectors (shown in green of 
Figure~\ref{fig:encoding}) and let
smoothing factor $\beta \rightarrow \infty$. With these modifications, 
the Encoding Layer acts like K-means. 
The Encoding Layer can also be regarded as a simplified version of GMM,
that allows different scaling (smoothing) of the clusters.
%The Encoding-layer can be extended to generalize GMM as discussed in Appendix~\ref{sec:appB}.

\paragraph{\bf Relation to BoWs and Residual Encoders}

BoWs (bag-of-word) methods typically   
hard assign each descriptor to the
nearest codeword  
and counts the occurrence of the 
visual words by 
aggregating the assignment vectors $\sum_{i=1}^N a_i$~\cite{shotton2006textonboost}. 
An improved BoW employs a soft-assignment weights~\cite{liu2011defense}.
VLAD~\cite{jegou2010aggregating} aggregates the residual vector with the hard-assignment 
weights.
NetVLAD \cite{jegou2010aggregating}  makes two relaxations:
(1) soft-assignment to make the model differentiable
and (2) decoupling the assignment from the dictionary 
which makes the assigning weights depend only on the input instead of the
dictionary.
% Hang, what does this mean "decouples the assignment from the dictionary". Another sentence is needed for clarification
% --as added, 'makes assigning weights independent with the dictionary'
%A key difference from our work is that
Therefore, the codewords 
are not learned from the distribution of the descriptors. Considering Figure~\ref{fig:encoding}, NetVLAD  drops
the link between visual words with their assignments (the blue 
arrow in Figure~\ref{fig:encoding}).
Fisher Vector~\cite{perronnin2010improving} concatenates both the 1st order and 2nd order aggregated 
residuals. FV-CNN~\cite{cimpoi15} encodes off-the-shelf CNNs with
pre-trained CNN and achieves good result in material recognition.
Fisher Kernel SVM~\cite{sydorov2014deep} iteratively
update the SVM by a convex solver and
the inner GMM parameters using gradient descent.
A key difference from our work is that this Fisher Kernel method uses hand-crafted instead of  learning
the features.
VLAD-CNN~\cite{gong2014multi} and FV-CNN~\cite{cimpoi15} build
off-the-shelf residual encoders with pre-trained CNNs and achieve 
great success in robust visual recognition and understanding areas.

\paragraph{\bf Relation to Pooling}
In CNNs, a pooling layer (Max or Avg) is typically used on top of the 
convolutional layers. 
Letting $K=1$ and fixing $c=0$, 
the Encoding Layer  simplifies to Sum pooling 
($e=\sum_{i=1}^N x_i$ and $\frac{d_\ell}{d_{x_i}}=\frac{d_\ell}{d_{e}}$). 
When followed by $L2$-normalization, 
it has exactly the same behavior as Avg pooling.
The convolutional layers extract features
as a sliding window, which can accept arbitrary input image sizes. 
However, the pooling layers 
usually have fixed receptive field size, which lead to the CNNs only
allowing fixed input image size.
SPP pooling layer~\cite{he2014spatial} accepts different size
by fixing the pooling bin number 
instead of receptive field sizes.
%which acts as resizing manner.
The relative spatial orders of the descriptors are preserved. 
Bilinear pooling layer~\cite{lin2015bilinear} 
removes the globally ordered information by
summing the outer-product of the descriptors across different locations.
Our Encoding Layer acts as a pooling layer by encoding robust residual
representations, which converts arbitrary input size to a fix length 
representation. 
Table~\ref{tab:compare} summarizes the comparison  our approach to other methods. 
%This creates a new opportunity of the neural network architectures.

\section{Deep Texture Encoding Network}
\label{sec:ten}
We refer to the deep convolutional neural network with the Encoding Layer as
{\it Deep Texture Encoding Network (Deep-TEN)}. 
In this section, we discuss the properties of the 
Deep-TEN, that is the property of integrating Encoding Layer with
an end-to-end CNN architecture.

\newcommand{\blocka}[2]{\multirow{3}{*}{\(\left[\begin{array}{c}\text{3$\times$3, #1}\\[-.1em] \text{3$\times$3, #1} \end{array}\right]\)$\times$#2}
}
\newcommand{\blockb}[3]{\multirow{3}{*}{\(\left[\begin{array}{c}\text{1$\times$1, #2}\\[-.1em] \text{3$\times$3, #2}\\[-.1em] \text{1$\times$1, #1}\end{array}\right]\)$\times$#3}
}
\renewcommand\arraystretch{1.1}
\setlength{\tabcolsep}{3pt}
\begin{table}[t]
\centering
\begin{adjustbox}{max width=0.5\textwidth}
\begin{tabular}{|l|c|c|c|c|c|c|c|c|}
 \hline
  & output size & Deep-TEN 50 \\
 \hline
 Conv1 & 176$\times$176$\times$64 & 7$\times$7, stride 2\\
  \hline
  \multirow{4}{*}{Res1} & \multirow{4}{*}{88$\times$88$\times$256} &     $3\times 3$ max pool, stride 2 \\\cline{3-3}
  &  & \blockb{256}{64}{3} \\
  & & \\
  & & \\\hline
  \multirow{3}{*}{Res2} & \multirow{3}{*}{44$\times$44$\times$512} 
  & \blockb{512}{128}{4} \\
  & & \\
  & & \\\hline
  \multirow{3}{*}{Res3} & \multirow{3}{*}{22$\times$22$\times$1024} 
  & \blockb{1024}{256}{6} \\
  & & \\
  & & \\\hline
  \multirow{3}{*}{Res4} & \multirow{3}{*}{11$\times$11$\times$2048} 
  & \blockb{2048}{512}{3} \\
  & & \\
  & & \\\hline
 %\multirow{2}{*}{
 Projection & \multirow{2}{*}{121$\times$128} &
conv 1$\times$1, 2048$\Rightarrow$128  \\\cline{3-3}
 + Reshape& & W$\times$H$\times$D$\Rightarrow$N$\times$D \\ 
 \hline
 Encoding & 32$\times$128 & $32$ codewords \\
 \hline
 $L2$-norm + FC & n classes & 1$\times$1 FC \\
 \hline
\end{tabular}
\end{adjustbox}
\caption{Deep-TEN architectures for adopting 50 layer pre-trained ResNet. 
The 2\textsuperscript{nd} column shows the featuremap sizes for input 
image size of 352$\times$352. When multi-size training for
input image size 320$\times$320,
the featuremap after Res4 is 10$\times$10.
We adopt a 1$\times$1 convolutional
layer after Res4 to reduce number of channels.
}
\label{tab:encoding-net}
\end{table}

\paragraph{\bf Domain Transfer}
% TODO compare with NetVLAD
Fisher Vector (FV) has the property of discarding the influence of
frequently appearing features in the dataset~\cite{perronnin2010improving}, 
which usually contains domain specific information
\cite{zhang16}. FV-CNN has shown its domain transfer ability practically
in material recognition work~\cite{cimpoi15}.
Deep-TEN generalizes the residual encoder and also preserves this 
property. To see this intuitively, consider the following: when a visual descriptor $x_i$
appears frequently in the data, it is likely to be close to 
one of the visual centers $c_k$. 
Therefore, the resulting residual vector corresponding to $c_k$,
$r_{ik} = x_i - c_k$, is small. 
For the residual vectors of $r_{ij}$ corresponding to $c_j$ 
where $j\ne k$,
the corresponding assigning weight
$a_{ij}$ becomes small as shown in Equation~\ref{eq:assigning}. 
The Encoding Layer aggregates the residual vectors with
assignment weights and results in small values for
frequently appearing visual descriptors.
This property is essential for transferring features
learned from different domain, and in this work we transfer
CNNs pre-trained on the object dataset ImageNet
to material recognition tasks.

% TODO FIXME, add detail
Traditional approaches do not have domain transfer problems because the features are usually generic and the domain-specific information is carried by the dictionary and encoding representations. 
The proposed Encoding Layer generalizes the dictionary learning and encoding framework, which carries domain-specific information. 
Because the entire network is optimized as a regression progress, the resulting convolutional features (with Encoding Layer learned on top) are likely to be domain-independent and therefore easier to transfer. 

% Hang, I removed the next sentence we are not experts in brain science and the reviewers will pounce
%Comparing with human brain, the convolutional features are like human optic nerves, which capturing any appearance in the world and the encoding layer learns the abstract thinking to a specific tasks.

\paragraph{Multi-size Training}
CNNs typically require a fixed input image size. 
In order to feed into the
network, images have to be resized or cropped to a fixed size. 
The convolutional layers act as
in sliding window manner, which can allow any input sizes 
(as discussed in SPP\cite{he2014spatial}). 
The FC (fully connected) layer acts as a classifier which take a fix length representation
as input. Our Encoding Layer act as a pooling layer
on top of the convolutional
layers, which converts arbitrary input sizes to a fixed 
length representation.
Our experiments show that the classification results are often improved by
iteratively training the Deep Encoding Network with different 
image sizes. In addition, this multi-size training provides
the opportunity for cross dataset training.

\paragraph{\bf Joint Deep Encoding}
There are many labeled datasets for different visual problems,
such as object classification~\cite{imagenet,cifar,coates2010analysis}, 
scene understanding~\cite{xiao2010sun,yu2015lsun}, 
object detection~\cite{pascal-voc-2007,lin2014microsoft}
and material recognition\cite{bell15,zhang15}.  
% Deep learning has achieved great success in those challenges.
An interesting question to ask is: how can different visual
tasks benefit each other? 
Different datasets have different domains, different labeling strategies
and sometimes different image sizes (e.g. 
 CIFAR10\cite{cifar} and ImageNet\cite{imagenet}).
Sharing convolutional features typically achieves great success~\cite{he2014spatial,ren2015faster}.
The concept of
{\it multi-task learning}~\cite{simonyan2014two} was originally proposed in~\cite{collobert2008unified},
to jointly train cross different datasets.
An issue in joint training is that features from different datasets
may not benefit from the combined training since the images contain
domain-specific information.
Furthermore, it is  typically not possible to learn deep features from
different image sizes.
Our Encoding Layer on top of convolution layers
accepts arbitrary input image sizes and learns
domain independent convolutional features, enabling convenient joint training. 
%and force the domain specific information in the encoding representation.
We present and evaluate a network that shares convolutional features
for two different dataset and has two separate Encoding Layers.
We demonstrate joint training with two datasets and 
show that recognition results are significantly improved.

\section{Experimental Results}

\begin{table*}[t]
\centering
\begin{adjustbox}{max width=0.95\textwidth}
\begin{tabular}{|l|c|c|c|c|c|c|c|c|}
 \hline
   & MINC-2500 & FMD &  GTOS & KTH & 4D-Light & MIT-Indoor & Caltech-101\\
 \hline
 FV-SIFT & 46.0 & 47.0 & 65.5 & 66.3 & 58.4 & 51.6 & 63.4 \\
 \hline
 % FMD: 76, 75, 75
 FV-CNN \small{(VGG-VD)} & 61.8 & 75.0 & 77.1 & 71.0 & 70.4 & 67.8 & 83.0\\  
 \hline
% FV-CNN \small{(Res50)} & coming & on & Monday \\ 
% \hline
% -------------------------------------------------------------
 % FMD: 80, 81, 79, 81, 80
 % MINC: 80.6 (s1)
 % GTOS 81.56(s1) 84.53(s2) 85.22(s4) 85.82(s5)
 % KTH: 85.0(s1) 84.4(s2) 77.9(s3) 80.6(s4)
 % 4D: 82.8(s1) 80.5(s2)  81.3(s3) 82.3(s5) (s4 have troubles)
 Deep-TEN {\small (\TextRed{ours})} & \textbf{80.6} & \textbf{80.2{\tiny$\pm 0.9$}} & \textbf{84.3{\tiny$\pm$1.9}} 
 & \textbf{82.0{\tiny$\pm$3.3}} & \textbf{81.7{\tiny$\pm$1.0}} & \textbf{71.3}  & \textbf{85.3} \\
 \hline
\end{tabular}
\end{adjustbox}
\caption{
The table compares the recognition results
of Deep-TEN with off-the-shelf encoding approaches, 
including Fisher Vector encoding of dense SIFT features (FV-SIFT) and
pre-trained CNN activations (FV-CNN) on different datasets
using single-size training.
Top-1 test accuracy mean$\pm$std \% is reported and the best 
result for each dataset is marked bold. 
{\small (The results of Deep-TEN for FMD, GTOS, KTH datasets 
are based on 5-time statistics,
and the results for MINC-2500, MIT-Indoor and Caltech-101
datasets are averaged over 2 runs. 
The baseline approaches are based on 1-time run.)}
}
\label{tab:benchmark}
\end{table*}

\begin{table*}[t]
\centering
\begin{adjustbox}{max width=0.95\textwidth}
\begin{tabular}{|l|c|c|c|c|c|c|c|c|}
%\Xhline{2\arrayrulewidth}
\hline
   & MINC-2500 & FMD &  GTOS & KTH & 4D-Light & MIT-Indoor \\
 \hline
 FV-CNN {\small(VGG-VD) multi} & 63.1 & 74.0 & 79.2 & 77.8 & 76.5 & 67.0   \\%\textbf{85.9} \\  
 \hline
  FV-CNN {\small(ResNet) multi} & 69.3 &78.2 & 77.1 & 78.3 & 77.6 & 76.1 \\
  \hline
  Deep-TEN {\small (\TextRed{ours})} & 80.6 &
  % \textbf{80.5} & \textbf{85.8} 
 \textbf{80.2{\tiny$\pm 0.9$}} & 84.3{\tiny$\pm$1.9}
 & 82.0{\tiny$\pm$3.3} & \textbf{81.7{\tiny$\pm$1.0}} & 71.3 \\
 \hline
 % -------------------------------------------------------------
 % GTOS: 80.38(s1) 85.44(s2) 84.32(s3) 88.77(s4) 84.87(s5)
 % KTH: 85.4 (s1) 88.6 (s2) 82.5(s3) 80.5(s4)
 % FMD 78, 78, 79, 79, 80
 % 4D: 82.5(s1) 79.1(s2) 79.4(s3) 84.7(s5) (s4 have troubles)
Deep-TEN {\small(\TextRed{ours}) multi} & \textbf{81.3}  & 78.8{\tiny$\pm$0.8} & \textbf{84.5{\tiny $\pm$2.9}} & \textbf{84.5{\tiny $\pm$3.5}} & 81.4{\tiny$\pm$2.6} & \textbf{76.2} \\%& - \\
 \hline
% -------------------------------------------------------------
 % ResNet50+SVM :
 % MIT:70.9
 %-------------------------------------------------------------
 % FMD: 
 %Deep-TEN {\small(\TextRed{ours})+SVM} &  & 81 & & & 76.4 &\\
%  \hline
%  State-of-the-Art & 76.0{\tiny$\pm$0.2}~\cite{bell15} & 
%  82.4{\tiny$\pm$1.4}~\cite{cimpoi15} & 
%  N/A& 
%  81.1{\tiny$\pm$1.5} \cite{cimpoi16} & 
%  77.0{\tiny $\pm$1.1}~\cite{wang2016dataset} &
%  80.0~\cite{cimpoi16}  &  91.44{\tiny$\pm$0.7}~\cite{he2014spatial}
%   \\ 
% \hline
\end{tabular}
\end{adjustbox}
\caption{Comparison of single-size and multi-size training. }
\label{tab:benchmark2}
\end{table*}

\begin{table*}[t]
\centering
\begin{adjustbox}{max width=0.95\textwidth}
\begin{tabular}{|l|c|c|c|c|c|c|c|c|}
\hline
   & MINC-2500 & FMD & GTOS & KTH & 4D-Light \\
  \hline
%  Deep-TEN {\small (\TextRed{ours})} & 80.6 & \textbf{80.5} & \textbf{85.8} 
%  & 82.0{\tiny$\pm$3.3} & \textbf{81.7{\tiny$\pm$1.0}}  \\
%  \hline
% Deep-TEN {\small(\TextRed{ours}) multi} & \textbf{81.3}  & 78.8{\tiny$\pm$0.8} & 84.5{\tiny $\pm$2.9} & \textbf{84.5{\tiny $\pm$3.5}} & 81.4{\tiny$\pm$2.6} \\
%  \hline
 Deep-TEN* {\small (\TextRed{ours})} & \textbf{81.3} & 80.2{\tiny$\pm$0.9} 
 & \textbf{84.5{\tiny$\pm$2.9}}
 & \textbf{84.5{\tiny$\pm$3.5}} & \textbf{81.7{\tiny$\pm$1.0}}  \\
 \hline
 State-of-the-Art & 76.0{\tiny$\pm$0.2}~\cite{bell15} & 
 \textbf{82.4{\tiny$\pm$1.4}}~\cite{cimpoi15} & 
 N/A& 
 81.1{\tiny$\pm$1.5} \cite{cimpoi16} & 
 77.0{\tiny $\pm$1.1}~\cite{wang2016dataset} 
 %& 80.0~\cite{cimpoi16}
 %&  91.44{\tiny$\pm$0.7}~\cite{he2014spatial}
  \\ 
 \hline
\end{tabular}
\end{adjustbox}
\caption{Comparison with state-of-the-art on four material/textures
dataset (GTOS is a new dataset, so  SoA is not available). 
Deep-TEN* denotes the best model of Deep Ten and Deep Ten {\small multi}. }
\label{tab:benchmark3}
\end{table*}

\paragraph{\bf Datasets}
The evaluation considers five material and texture datasets.
{\it Materials in Context Database} (MINC)~\cite{bell15} is a large scale material in the wild dataset. In this work, a publicly available subset (MINC-2500, Sec 5.4 of original paper) is evaluated with provided train-test splits, containing 23 material categories and 2,500 images per-category.
{\it Flickr Material Dataset} (FMD)~\cite{Sharan13}, a popular benchmark for material recognition containing 10 material classes, 90 images per-class used for training and 10 for test.
{\it Ground Terrain in Outdoor Scenes Dataset} (GTOS)~\cite{xue16} is a dataset of ground materials in outdoor scene with 40 categories. The evaluation is based on provided train-test splits.
{\it KTH-TIPS-2b} (KTH)-\cite{caputo2005class}, contains 11 texture categories and four samples per-category. Two samples are randomly picked for training and the others for test.
{\it 4D-Light-Field-Material} (4D-Light)~\cite{wang2016dataset} is a recent light-field material dataset containing 12 material categories with 100 samples per-category. In this experiment, 70 randomly picked samples 
per-category are used as training and the others for test and only one angular resolution is used per-sample.
For general classification evaluations,
two additional datasets are considered.
{\it MIT-Indoor}~\cite{quattoni2009recognizing} dataset is an indoor scene categorization dataset 
with 67 categories,
a standard subset of 80 images per-category for training and 20
for test is used in this work.
{\it Caltech 101}~\cite{fei2007learning} is a 102 category
(1 for background) 
object classification dataset; 10\% randomly picked samples are used for
test and the others for training.

\begin{figure}[t]
\begin{center}
\includegraphics[width=0.8\linewidth]{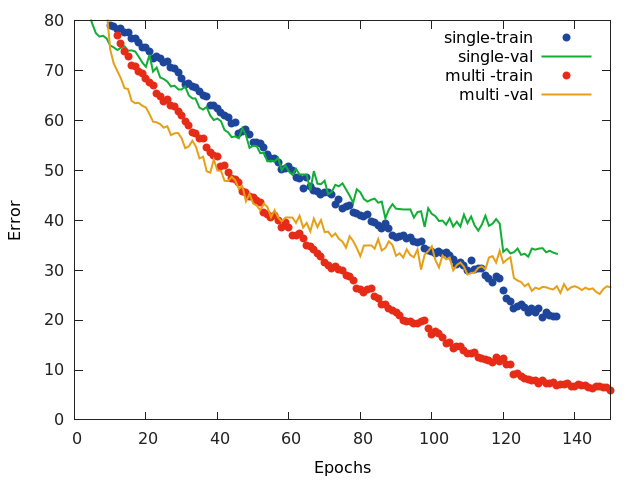}
\includegraphics[width=0.8\linewidth]{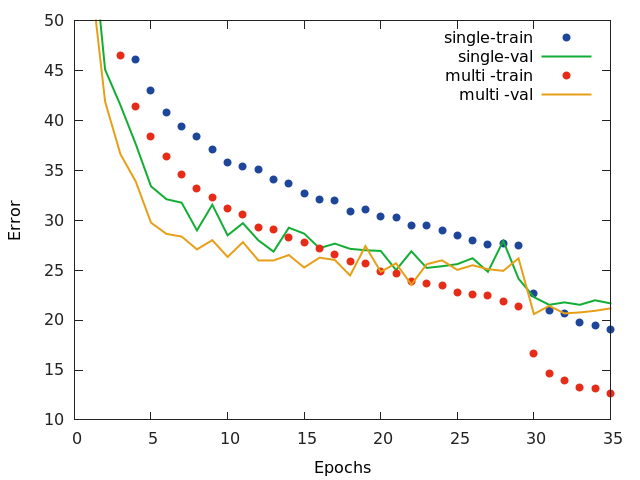}
\end{center}
\caption{ 
Comparison between single-size training and multi-size training. 
Iteratively training the Deep-TEN with two different intput sizes
(352$\times$352 and 320$\times$320) makes the network converging 
faster and improves the performance.
The top figure shows the training curve on MIT-Indoor 
 and the bottom one shows the first 35 epochs on MINC-2500.
}
\label{fig:multisize}
\end{figure}

\paragraph{\bf Baselines}
In order to evaluate different encoding and representations,
we  benchmark different approaches
with single input image sizes without ensembles, 
since we expect that the performance is likely to improve
by assembling features or using multiple scales.
We fix the input image size to 352$\times$352 for SIFT, pre-trained
CNNs feature extractions and Deep-TEN.
{FV-SIFT}, a non-CNN approach, is considered due to its similar 
encoding representations.  SIFT features of 128 dimensions are extracted
from input images and a GMM of 128 Gaussian components is built, resulting in a 
32K Fisher Vector encoding. 
For {FV-CNN} encoding, the CNN features of input images 
are extracted using pre-trained 16-layer VGG-VD 
model~\cite{simonyan2014very}.
The feature maps of conv5 (after ReLU) are used, 
with the dimensionality of 14$\times$14$\times$512.
Then a GMM of 32 Gaussian components is built and resulting in
a 32K FV-CNN encoding. 
To improve the results further, we build a stronger baseline using
pre-trained 50-layers ResNet~\cite{he2015deep} features.
The feature maps of the last residual unit are used. The extracted
features are projected into 512 dimension using PCA, from 
the large channel numbers of 2048 in ResNet. 
Then we follow the same encoding approach of standard FV-CNN to build
with ResNet features.
For comparison with multi-size training Deep-TEN, multi-size 
FV-CNN (VD) is used, the CNN features are extracted from two
different sizes of input image, 352$\times$352 and 320$\times$320 (sizes determined empirically). 
All the baseline encoding representations are reduced to 4096 
dimension using PCA and $L2$-normalized.
For classification, linear one-vs-all
Support Vector Machines (SVM) are built using the off-the-shelf
representations. The learning hyper-parameter is set to $C_{svm}=1$,
since the features are $L2$-normalized. 
The trained SVM classifiers
are recalibrated as in prior work~\cite{cimpoi15,lin2015bilinear}, 
by scaling the weights and biases such that the 
median prediction score of positive and negative samples are at $+1$ 
and $-1$.

\paragraph{Deep-TEN Details}
% network architecture
We build Deep-TEN with the architecture of an 
Encoding Layer on top of 50-layer pre-trained ResNet 
(as shown in Table~\ref{tab:encoding-net}). 
Due to high-dimensionality of 
ResNet feature maps on Res4, a $1\times 1$ convolutional layer is
used for reducing number of channels (2048$\Rightarrow$128).
Then an Encoding Layer with 32  codewords
is added on top, followed by $L2$-normalization and FC layer. 
The weights (codewords $C$ and smoothing factor $s$) are randomly initialized
with uniform distribution $\pm \frac{1}{\sqrt[]{K}}$. 
%For training from scratch, we adopt a smaller network architecture, dividing the number of featuremap channels of original ResNet-50 by 4 (as shown in Table~\ref{tab:encoding-net}).
For data augmentation, the input images are resized to 
400 along the short-edge with the per-pixel mean subtracted.
For in-the-wild image database,
the images are randomly cropped to 
9$\%$ to 100$\%$ of the image areas, keeping the aspect
ratio between 3$/$4 and 4$/$3. 
For the material database with in-lab or controlled conditions 
(KTH or GTOS), we keep the original image scale.
The resulting images are then resized into 352$\times$352 for 
single-size training (and 320$\times$320 for multi-size training),
with 50$\%$ chance horizontal flips.
Standard color augmentation is used as 
in~\cite{krizhevsky2012imagenet}. 
We use SGD with a mini-batch size of 64. 
%For training from scratch, the learning rate starts from 0.1 and is divided by 10 and 100 at 80\textsuperscript{th}  and 120\textsuperscript{th} epochs, and the models are trained for up to 160 epochs. 
For fine-tuning, the learning rate starts from 0.01 and
divided by 10 when the error plateaus.
We use a weight decay of 0.0001 and a momentum of 0.9.
In testing, we adopt standard 
10-crops~\cite{krizhevsky2012imagenet}. 

\paragraph{Multi-size Training}
Deep-TEN ideally can accept arbitrarily input image sizes (larger than
a constant).
In order to learn the network without modifying the standard  
optimization solver,
we train the network with a pre-defined size in each epoch and 
iteratively change the input image size for every epoch as 
in~\cite{he2014spatial}. 
A full evaluation of combinatorics of different size pairs have not
yet been explored.
Empirically, we consider two different sizes 
352$\times$352 and 320$\times$320 during the training and only
use single image size in testing for simplicity (352$\times$352). 
The two input
sizes result in 11$\times$11 and 10$\times$10 feature map sizes 
before feeding into the Encoding Layer.
Our goal is to evaluate how multi-size training
affects the network optimization and how the multi-scale features
affect texture recognition.

\subsection{Recognition Results} 
We evaluate the performance of
Deep-TEN, FV-SIFT and FV-CNN
on aforementioned golden-standard material and 
texture datasets, such as MINC-2500, FMD, KTH and two new material
datasets: 4D-Light and GTOS.  Additionally, two general recognition 
datasets MIT-Indoor and Caltech-101
are also considered.
%We benchmark different approaches with both single-size and
%multi-size training.
Table~\ref{tab:benchmark} shows overall experimental results using single-size training,

\paragraph{\bf Comparing with Baselines} 
% General description
As shown in Table~\ref{tab:benchmark}, Deep-TEN and FV-CNN always outperform FV-SIFT, which shows that
pre-trained CNN features are typically more discriminant
than hand-engineered SIFT features.
FV-CNN usually achieves reasonably good results  on different datasets
without fine-tuning pre-trained features.
We can observe that the 
performance of FV-CNN is often improved by employing ResNet features 
comparing with VGG-VD as shown in Table~\ref{tab:benchmark2}.
Deep-TEN outperforms FV-CNN under the same settings, which
shows that the Encoding Layer gives the advantage of transferring
pre-trained features to material recognition
by removing domain-specific information 
as described in Section~\ref{sec:ten}. 
% Talk about layer property
The  Encoding Layer's  property of representing feature distributions
is especially good for texture understanding and segmented material 
recognition. Therefore, Deep-TEN works well on GTOS and KTH datasets.
% Special case analysis
For the small-scale dataset FMD with less
training sample variety, Deep-TEN still outperforms the baseline
approaches that use an SVM classifier.
For MINC-2500, a relatively large-scale dataset, 
the end-to-end framework of Deep TEN shows its distinct advantage of
optimizing CNN features 
and consequently, the recognition results are significantly improved 
(61.8\%$\Rightarrow$80.6\% and 69.3\%$\Rightarrow$81.3,
compared with off-the-shelf representation of FV-CNN). 
% Feature distribution for scene understanding
For the MIT-Indoor dataset, the Encoding Layer
 works well on scene categorization
due to the need for a certain level of orderless and invariance. 
The best performance of these methods for  Caltech-101 is 
achieved by FV-CNN(VD) {\small multi} (85.7\% omitted from the table). 
The CNN models VGG-VD and ResNet are pre-trained on
ImageNet, which is also an object classification dataset like Caltech-101. The 
pre-trained features are discriminant to target datasets.
Therefore, Deep-TEN performance is only slightly better than the off-the-shelf representation FV-CNN.

%Multi-size Figure~\ref{fig:multisize} also shows that
\paragraph{\bf Impact of Multi-size} 
For in-the-wild datasets,
such as MINC-2500 and MIT-Indoor, 
the performance of all the approaches are improved by
adopting multi-size as expected. 
Remarkably,  as shown in Table~\ref{tab:benchmark2},
Deep-TEN shows a performance boost of 4.9\% 
using multi-size training and outperforms the best baseline
by 7.4\% on MIT-Indoor dataset. 
For 
some 
%smaller scale or in-controlled condition 
datasets such as FMD and GTOS, 
the performance decreases slighltly by adopting 
multi-size training due to lack of variety in the training data.
Figure~\ref{fig:multisize} compares the single-size training and
multi-size (two-size) training for Deep-TEN
on MIT-Indoor and MINC-2500 dataset.
The experiments show  that
multi-size training helps the optimization of the network
(converging faster) and
the learned multi-scale features are useful for
the recognition. % even with single-scale testing.

%The validation error got less improved than the training, because single-size test for Deep-TEN is adopted in this experiment.

%You've addressed this to some degree saying that the method works best with large scale data. I think that you should say end-to-end training provides additional advantages (can be improved by adding more data, does not need manual tuning.)

\paragraph{\bf Comparison with State-of-the-Art} 
As shown in Table~\ref{tab:benchmark3}, Deep-TEN outperforms the state-of-the-art on four material/texture recognition 
datasets: MINC-2500, KTH, GTOS and 4D-Light. 
Deep-TEN also performs well on two general
recognition datasets. 
Notably, the prior state-of-the-art
approaches either (1) relies on assembling features (such as FV-SIFT \& CNNs)
and/or (2) adopts an additional SVM classifier for classification. 
Deep-TEN as an end-to-end framework neither concatenates
any additional hand-engineered features nor employe SVM
for classification.
For the small-scale datasets such as FMD and MIT-Indoor (subset),
the proposed Deep-TEN gets compatible results with 
state-of-the-art approaches (FMD within 2\%, MIT-indoor within 4\%). 
For the large-scale datasets such as MINC-2500,
%and GTOS datasets,
Deep-TEN outperforms the prior work and baselines by a large margin
demonstrating its great advantage of end-to-end learning and the ability
of transferring pre-trained CNNs. 
We expect that the performance of Deep-TEN can scale better than
traditional approaches when adding 
more training data. 

\subsection{Joint Encoding from Scratch}

\begin{table}[t]
\centering
\label{tab:joint}
\begin{adjustbox}{max width=0.45\textwidth}
\begin{tabular}{|l|c|c|c|c|}
 \hline
  & STL-10 & CIFAR-10 \\
 \hline
 Deep-TEN \small{(Individual)} & 76.29 & 91.5 \\
 \hline
 Deep-TEN \small{(Joint)} & \textbf{87.11} & 91.8 \\
 \hline
 State-of-the-Art & 74.33~\cite{zhao2015stacked} & -  \\
 \hline
\end{tabular}
\end{adjustbox}
\caption{Joint Encoding on CIFAR-10 and STL-10 datasets. 
Top-1 test accuracy \%. When joint training with CIFAR-10,
the recognition result on STL-10 got significantly improved. 
{\small (Note that traditional network architecture does not allow joint
training with different image sizes.)}}
\end{table}

We test joint training on two small datasets 
CIFAR-10~\cite{cifar}
and STL-10~\cite{coates2010analysis}  
as a litmus test of Joint Encoding from scratch. 
We expect the convolutional features learned with Encoding Layer
are easier to transfer, 
and can improve the recognition on both datasets. 

CIFAR-10 contains 60,000 tiny images with the size 32$\times$32 belonging 
to 10 classes
(50,000 for training and 10,000 for test),
which is a subset of tiny images database.
STL-10 is a dataset acquired from ImageNet~\cite{imagenet} and
originally designed for unsupervised feature learning, which
has 5,000 labeled images for training and 8,000 for test with the 
size of $96\times96$. 
For the STL-10 dataset only the labeled images are used for training. 
Therefore, learning CNN from scratch is not
supposed to work well due to the limited training data.
We make a very simple network architecture,
by simply replacing the $8\times8$ Avg pooling layer of pre-Activation
ResNet-20\cite{he2016identity} with 
%$24\times24$ Avg pooling (baseline) or 
Encoding-Layer (16 codewords). 
We then build a network with shared convolutional layers and 
separate encoding layers that is jointly trained on two datasets.
Note that the traditional CNN architecture is not applicable due to 
different image sizes from this two datasets. 
The training loss is computed as the sum of
the two classification losses, 
and the gradient of the convolutional layers
are accumulated together.
For data augmentation
in the training: 4 pixels are padded on each side for CIFAR-10
and 12 pixels for STL-10,
and then randomly crop the padded images or its horizontal flip
into original sizes 32$\times$32 for CIFAR-10 and 96$\times$96 for
STL-10. 
For testing, we only evaluate
the single view of the original images.
The model is trained with a mini batch of 128 for each dataset.
We start with a learning
rate of 0.1 and divide it by 10 and 100 at 80\textsuperscript{th} and 
120\textsuperscript{th} epoch.

The experimental results show that the recognition result of
STL-10 dataset is significantly improved by joint training
the Deep TEN with CIFAR-10 dataset. Our approach 
achieves the recognition
rate of 87.11$\%$, which outperforms previous 
the state of the art 74.33$\%$~\cite{zhao2015stacked}
and 72.8$\%$\cite{dosovitskiy2014discriminative}
by a large margin.

\section{Conclusion}
In summary,   
we developed a Encoding Layer which bridges the gap between classic computer vision approaches and the CNN architecture, (1) making the deep learning framework more flexible by allowing arbitrary input image size,
(2) making the learned convolutional features easier to transfer since the Encoding Layer is likely to carry domain-specific information. 
The Encoding Layer shows superior performance of transferring pre-trained CNN features. 
Deep-TEN outperforms traditional off-the-shelf methods and achieves state-of-the-art results on MINC-2500, KTH and two recent material datasets: GTOS and 4D-Lightfield. 
%We demonstrated the effectiveness on various material and texture recognition datasets.
%In addition, the experimental results for joint encoding on two different datasets are very promising, which provides a new opportunity for different vision tasks to mutually benefit  each other.

The Encoding Layer is efficient using GPU computations and
our Torch~\cite{torch7} implementation
of 50-layer Deep-Ten (as shown in Table~\ref{tab:encoding-net}) 
takes the input images
of size 352$\times$352, runs at 55 frame/sec for training and 290
frame/sec for inference on 4 Titan X Maxwell GPUs.

\section*{Acknowledgment}
This work was supported by National Science Foundation 
award IIS-1421134. A GPU used for this research was 
donated by the NVIDIA Corporation.

\appendix
\section{Encoding Layer Implementations}
\label{sec:appA}
This appendix section provides 
the explicit expression for the gradients of the loss $\ell$ 
with respect to ({\it w.r.t}) the layer input and the parameters 
for implementing Encoding Layer.
The $L2$-normalization as a standard component
is used outside the encoding layer.

\paragraph{\bf Gradients {w.r.t} Input $X$} 
The encoder $E=\{e_1,...e_K\}$ can be viewed as $k$ independent sub-encoders.
Therefore the gradients of the loss function $\ell$ {\it w.r.t}
input descriptor $x_i$ can be accumulated $\frac{d_\ell}{d_{x_i}} = \sum_{k=1}^K \frac{d_\ell}{d_{e_{k}}}\cdot \frac{d_{e_{k}}}{d_{x_i}}$. 
According to the chain rule, the gradients of the encoder
$w.r.t$ the input is given by
\begin{equation}
\frac{d_{e_{k}}}{d_{x_{i}}}
=r_{ik}^T \frac{d_{a_{ik}}}{d_{x_{i}}} + 
a_{ik} \frac{d_{r_{ik}}}{d_{x_{i}}} ,
\end{equation}
where $a_{ik}$ and $r_{ik}$ are defined in Sec~\ref{sec:encoding}, $\frac{d_{r_{ik}}}{d_{x_i}} = 1$. 
Let $f_{ik}=e^{-s_k \| r_{ik}\|^2}$ and $h_i=\sum_{m=1}^K f_{im}$,
 we can write $a_{ik} = \frac{f_{ik}}{h_i}$.
The derivatives of the assigning weight
{\it w.r.t} the input descriptor is 
\begin{equation}
\frac{d_{a_{ik}}}{d_{x_i}}=\frac{1}{h_i}\cdot \frac{d_{f_{ik}}}{d_{x_i}} - \frac{f_{ik}}{(h_i)^2}\cdot \sum_{m=1}^K \frac{d_{f_{im}}}{d_{x_i}} ,
\end{equation}
where $\frac{d_{f_{ik}}}{d_{x_i}}=-2s_k f_{ik}\cdot r_{ik}$.

\paragraph{\bf Gradients {w.r.t} Codewords $C$} 
The sub-encoder $e_k$ only depends on the codeword $c_k$. 
Therefore, the gradient of loss function {\it w.r.t} the codeword is given by 
$\frac{d_\ell}{d_{c_k}}=\frac{d_\ell}{d_{e_k}}\cdot \frac{d_{e_k}}{d_{c_k}}$. 
\begin{equation}
\begin{split}
	\frac{d_{e_k}}{d_{c_k}}&=\sum_{i=1}^N(r_{ik}^T \frac{d_{a_{ik}}}{d_{c_k}}+a_{ik}\frac{d_{r_{ik}}}{d_{c_k}}) ,\\
 %   &=\sum_{i=1}^N r_{ik}^T \frac{d_{a_{ik}}}{d_{c_k}} - \sum_{i=1}^N a_{ik}
\end{split}
\end{equation}
where $\frac{d_{r_{ik}}}{d_{c_k}} = -1$. 
Let  $g_{ik}=\sum_{m\ne k} f_{im}$.
According to the chain rule, the derivatives of assigning {\it w.r.t} the 
codewords can be written as
\begin{equation}
\frac{d_{a_{ik}}}{d_{c_k}}=\frac{d_{a_{ik}}}{d_{f_{ik}}}\cdot \frac{d_{f_{ik}}}{d_{c_k}}=\frac{2s_k f_{ik}g_{ik}}{(h_i)^2}\cdot r_{ik} .
\end{equation}

\paragraph{\bf Gradients {w.r.t} Smoothing Factors} 
Similar to the codewords, the sub-encoder $e_k$ only depends on
the $k$-th smoothing factor $s_k$. Then, the gradient of the loss function
{\it w.r.t} the smoothing weight is given by $\frac{d_\ell}
{d_{s_k}}=\frac{d_\ell}{d_{e_k}}\cdot \frac{d_{e_k}}{d_{s_k}}$.

\begin{equation}
\frac{d_{e_k}}{d_{s_k}}=-\frac{f_{ik}g_{ik}\|r_{ik}\|^2}{(h_i)^2}
\end{equation}

\paragraph{\bf Note} In practice, we multiply
the numerator and denominator 
of the assigning weight with $e^{\phi_i}$ to avoid overflow:
%as in the softmax layer:
\begin{equation}
a_{ik}=
\frac{\exp(-s_k \|r_{ik}\|^2+\phi_i)}{\sum_{j=1}^K\exp(-s_j \|r_{ij}\|^2+\phi_i)} ,
\end{equation}
where $\phi_i=\min_k\{s_k\|r_{ik}\|^2\}$.
Then $\frac{d_{\bar{f}_{ik}}}{d_{x_i}}=e^{\phi_i}\frac{f_{ik}}{d_{x_i}}$.

A Torch~\cite{torch7} implementation is provided 
in supplementary material and available at
\url{https://github.com/zhanghang1989/Deep-Encoding}.

\section{Multi-size Training-from-Scratch}
\begin{figure}
\includegraphics[width=\linewidth]{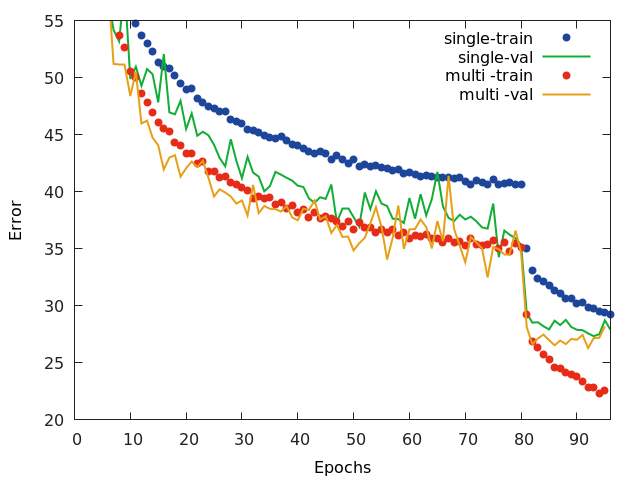}
\caption{Training Deep-TEN-50 from-scratch on MINC-2500 (first 100 epochs). 
Multi-size training helps the optimization of Deep-TEN
and improve the performance. %Notice the error rates for multi-size training (both the training and the validation errors) are  lower than the single-size. The dip near 80\textsuperscript{th} epoch is due to decreasing the learning rate.  
}
\label{fig:scratch}
\end{figure}

We also tried to train Deep-TEN from-scratch on MINC-2500
, 
the result is omitted in the main
paper due to having inferior recognition performance comparing with 
employing pre-trained ResNet-50. 
As shown in Figure~\ref{fig:scratch}, the converging 
speed is significantly improved using
multi-size training, which proves our hypothesis that
multi-size training helps the optimization of the network. 
The validation error is less improved than the training error, 
since we adopt single-size test for simplicity.

\begin{figure}[t]
\centering
\includegraphics[width=0.6\linewidth]{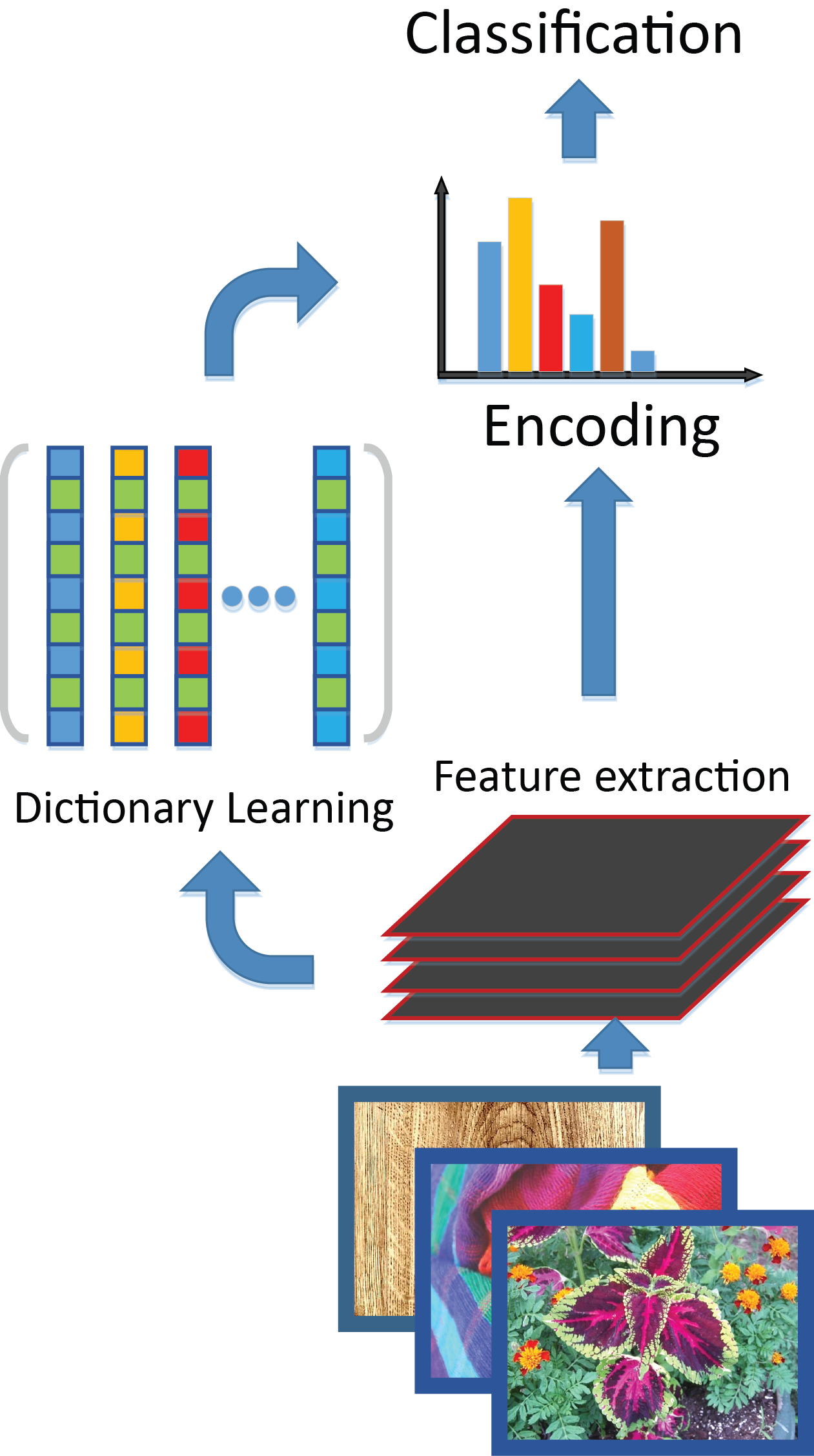}
\caption{
Pipelines of classic computer vision approaches. 
Given in put images, the local visual appearance is extracted using
hang-engineered features (SIFT or filter bank responses). A dictionary
is then learned off-line using unsupervised grouping such as K-means. 
An encoder (such as BoWs or Fisher Vector) is built on top which describes the
distribution of the features and output a fixed-length representations
for classification.
}
\label{fig:pipeline}
\end{figure}

\begin{comment}
\section{GMM Extension}
\label{sec:appB}
Gaussian Mixture Model (GMM) allows a finer distribution modeling,
which is represented as a weighted sum of several Gaussian components.  
This allows different distribution shapes of different clusters and also
allows different mixture weights for the clusters. 

\begin{quotation}
s
\end{quotation}

\paragraph{\bf Learnable Fisher Vector Encoding}

\end{comment}

{\small
\bibliographystyle{ieee}
\bibliography{cvpr17}

\begin{thebibliography}{10}\itemsep=-1pt

\bibitem{arandjelovic2015netvlad}
R.~Arandjelovi{\'c}, P.~Gronat, A.~Torii, T.~Pajdla, and J.~Sivic.
\newblock Netvlad: Cnn architecture for weakly supervised place recognition.
\newblock {\em arXiv preprint arXiv:1511.07247}, 2015.

\bibitem{bell15}
S.~Bell, P.~Upchurch, N.~Snavely, and K.~Bala.
\newblock Material recognition in the wild with the materials in context
  database.
\newblock In {\em Proceedings of the IEEE Conference on Computer Vision and
  Pattern Recognition}, pages 3479--3487, 2015.

\bibitem{caputo2005class}
B.~Caputo, E.~Hayman, and P.~Mallikarjuna.
\newblock Class-specific material categorisation.
\newblock In {\em Computer Vision, 2005. ICCV 2005. Tenth IEEE International
  Conference on}, volume~2, pages 1597--1604. IEEE, 2005.

\bibitem{cimpoi16}
M.~Cimpoi, S.~Maji, I.~Kokkinos, and A.~Vedaldi.
\newblock Deep filter banks for texture recognition, description, and
  segmentation.
\newblock {\em International Journal of Computer Vision}, 118(1):65--94, 2016.

\bibitem{cimpoi15}
M.~Cimpoi, S.~Maji, and A.~Vedaldi.
\newblock Deep filter banks for texture recognition and segmentation.
\newblock In {\em Proceedings of the IEEE Conference on Computer Vision and
  Pattern Recognition}, pages 3828--3836, 2015.

\bibitem{coates2010analysis}
A.~Coates, H.~Lee, and A.~Y. Ng.
\newblock An analysis of single-layer networks in unsupervised feature
  learning.
\newblock {\em Ann Arbor}, 1001(48109):2, 2010.

\bibitem{torch7}
R.~Collobert, K.~Kavukcuoglu, and C.~Farabet.
\newblock Torch7: A matlab-like environment for machine learning.
\newblock In {\em BigLearn, NIPS Workshop}, number EPFL-CONF-192376, 2011.

\bibitem{collobert2008unified}
R.~Collobert and J.~Weston.
\newblock A unified architecture for natural language processing: Deep neural
  networks with multitask learning.
\newblock In {\em Proceedings of the 25th international conference on Machine
  learning}, pages 160--167. ACM, 2008.

\bibitem{csurka2004visual}
G.~Csurka, C.~Dance, L.~Fan, J.~Willamowski, and C.~Bray.
\newblock Visual categorization with bags of keypoints.
\newblock In {\em Workshop on statistical learning in computer vision, ECCV},
  volume~1, pages 1--2. Prague, 2004.

\bibitem{Cula01b}
O.~G. Cula and K.~J. Dana.
\newblock Compact representation of bidirectional texture functions.
\newblock {\em IEEE Conference on Computer Vision and Pattern Recognition},
  1:1041--1067, December 2001.

\bibitem{Cula01a}
O.~G. Cula and K.~J. Dana.
\newblock Recognition methods for 3d textured surfaces.
\newblock {\em Proceedings of SPIE Conference on Human Vision and Electronic
  Imaging VI}, 4299:209--220, January 2001.

\bibitem{imagenet}
J.~Deng, W.~Dong, R.~Socher, L.-J. Li, K.~Li, and L.~Fei-Fei.
\newblock {ImageNet: A Large-Scale Hierarchical Image Database}.
\newblock In {\em CVPR09}, 2009.

\bibitem{dosovitskiy2014discriminative}
A.~Dosovitskiy, J.~T. Springenberg, M.~Riedmiller, and T.~Brox.
\newblock Discriminative unsupervised feature learning with convolutional
  neural networks.
\newblock In {\em Advances in Neural Information Processing Systems}, pages
  766--774, 2014.

\bibitem{pascal-voc-2007}
M.~Everingham, L.~Van~Gool, C.~K.~I. Williams, J.~Winn, and A.~Zisserman.
\newblock The {PASCAL} {V}isual {O}bject {C}lasses {C}hallenge 2007 {(VOC2007)}
  {R}esults.
\newblock
  http://www.pascal-network.org/challenges/VOC/voc2007/workshop/index.html.

\bibitem{everitt1981finite}
B.~S. Everitt.
\newblock {\em Finite mixture distributions}.
\newblock Wiley Online Library, 1981.

\bibitem{fei2007learning}
L.~Fei-Fei, R.~Fergus, and P.~Perona.
\newblock Learning generative visual models from few training examples: An
  incremental bayesian approach tested on 101 object categories.
\newblock {\em Computer Vision and Image Understanding}, 106(1):59--70, 2007.

\bibitem{fei2005bayesian}
L.~Fei-Fei and P.~Perona.
\newblock A bayesian hierarchical model for learning natural scene categories.
\newblock In {\em 2005 IEEE Computer Society Conference on Computer Vision and
  Pattern Recognition (CVPR'05)}, volume~2, pages 524--531. IEEE, 2005.

\bibitem{gong2014multi}
Y.~Gong, L.~Wang, R.~Guo, and S.~Lazebnik.
\newblock Multi-scale orderless pooling of deep convolutional activation
  features.
\newblock In {\em European Conference on Computer Vision}, pages 392--407.
  Springer, 2014.

\bibitem{he2014spatial}
K.~He, X.~Zhang, S.~Ren, and J.~Sun.
\newblock Spatial pyramid pooling in deep convolutional networks for visual
  recognition.
\newblock In {\em European Conference on Computer Vision}, pages 346--361.
  Springer, 2014.

\bibitem{he2015deep}
K.~He, X.~Zhang, S.~Ren, and J.~Sun.
\newblock Deep residual learning for image recognition.
\newblock {\em arXiv preprint arXiv:1512.03385}, 2015.

\bibitem{he2016identity}
K.~He, X.~Zhang, S.~Ren, and J.~Sun.
\newblock Identity mappings in deep residual networks.
\newblock {\em arXiv preprint arXiv:1603.05027}, 2016.

\bibitem{jegou2010aggregating}
H.~J{\'e}gou, M.~Douze, C.~Schmid, and P.~P{\'e}rez.
\newblock Aggregating local descriptors into a compact image representation.
\newblock In {\em Computer Vision and Pattern Recognition (CVPR), 2010 IEEE
  Conference on}, pages 3304--3311. IEEE, 2010.

\bibitem{joachims1998text}
T.~Joachims.
\newblock Text categorization with support vector machines: Learning with many
  relevant features.
\newblock In {\em European conference on machine learning}, pages 137--142.
  Springer, 1998.

\bibitem{cifar}
A.~Krizhevsky and G.~Hinton.
\newblock Learning multiple layers of features from tiny images.
\newblock {\em University of Toronto, Technical Report}, 2009.

\bibitem{krizhevsky2012imagenet}
A.~Krizhevsky, I.~Sutskever, and G.~E. Hinton.
\newblock Imagenet classification with deep convolutional neural networks.
\newblock In {\em Advances in neural information processing systems}, pages
  1097--1105, 2012.

\bibitem{leung2001representing}
T.~Leung and J.~Malik.
\newblock Representing and recognizing the visual appearance of materials using
  three-dimensional textons.
\newblock {\em International journal of computer vision}, 43(1):29--44, 2001.

\bibitem{lin2014microsoft}
T.-Y. Lin, M.~Maire, S.~Belongie, J.~Hays, P.~Perona, D.~Ramanan,
  P.~Doll{\'a}r, and C.~L. Zitnick.
\newblock Microsoft coco: Common objects in context.
\newblock In {\em European Conference on Computer Vision}, pages 740--755.
  Springer, 2014.

\bibitem{lin2015bilinear}
T.-Y. Lin, A.~RoyChowdhury, and S.~Maji.
\newblock Bilinear cnn models for fine-grained visual recognition.
\newblock In {\em Proceedings of the IEEE International Conference on Computer
  Vision}, pages 1449--1457, 2015.

\bibitem{liu2011defense}
L.~Liu, L.~Wang, and X.~Liu.
\newblock In defense of soft-assignment coding.
\newblock In {\em 2011 International Conference on Computer Vision}, pages
  2486--2493. IEEE, 2011.

\bibitem{lloyd1982least}
S.~Lloyd.
\newblock Least squares quantization in pcm.
\newblock {\em IEEE transactions on information theory}, 28(2):129--137, 1982.

\bibitem{lowe2004distinctive}
D.~G. Lowe.
\newblock Distinctive image features from scale-invariant keypoints.
\newblock {\em International journal of computer vision}, 60(2):91--110, 2004.

\bibitem{perronnin2010improving}
F.~Perronnin, J.~S{\'a}nchez, and T.~Mensink.
\newblock Improving the fisher kernel for large-scale image classification.
\newblock In {\em European conference on computer vision}, pages 143--156.
  Springer, 2010.

\bibitem{quattoni2009recognizing}
A.~Quattoni and A.~Torralba.
\newblock Recognizing indoor scenes.
\newblock In {\em Computer Vision and Pattern Recognition, 2009. CVPR 2009.
  IEEE Conference on}, pages 413--420. IEEE, 2009.

\bibitem{ren2015faster}
S.~Ren, K.~He, R.~Girshick, and J.~Sun.
\newblock Faster r-cnn: Towards real-time object detection with region proposal
  networks.
\newblock In {\em Advances in neural information processing systems}, pages
  91--99, 2015.

\bibitem{Sharan13}
L.~Sharan, C.~Liu, R.~Rosenholtz, and E.~H. Adelson.
\newblock Recognizing materials using perceptually inspired features.
\newblock {\em International journal of computer vision}, 103(3):348--371,
  2013.

\bibitem{shotton2006textonboost}
J.~Shotton, J.~Winn, C.~Rother, and A.~Criminisi.
\newblock Textonboost: Joint appearance, shape and context modeling for
  multi-class object recognition and segmentation.
\newblock In {\em European conference on computer vision}, pages 1--15.
  Springer, 2006.

\bibitem{simonyan2014two}
K.~Simonyan and A.~Zisserman.
\newblock Two-stream convolutional networks for action recognition in videos.
\newblock In {\em Advances in Neural Information Processing Systems}, pages
  568--576, 2014.

\bibitem{simonyan2014very}
K.~Simonyan and A.~Zisserman.
\newblock Very deep convolutional networks for large-scale image recognition.
\newblock {\em arXiv preprint arXiv:1409.1556}, 2014.

\bibitem{sivic2005discovering}
J.~Sivic, B.~C. Russell, A.~A. Efros, A.~Zisserman, and W.~T. Freeman.
\newblock Discovering objects and their location in images.
\newblock In {\em Tenth IEEE International Conference on Computer Vision
  (ICCV'05) Volume 1}, volume~1, pages 370--377. IEEE, 2005.

\bibitem{sydorov2014deep}
V.~Sydorov, M.~Sakurada, and C.~H. Lampert.
\newblock Deep fisher kernels-end to end learning of the fisher kernel gmm
  parameters.
\newblock In {\em Proceedings of the IEEE Conference on Computer Vision and
  Pattern Recognition}, pages 1402--1409, 2014.

\bibitem{Gemert08}
J.~C. van Gemert, J.-M. Geusebroek, C.~J. Veenman, and A.~W.~M. Smeulders.
\newblock Kernel codebooks for scene categorization.
\newblock In {\em ECCV 2008, PART III. LNCS}, pages 696--709. Springer, 2008.

\bibitem{varma2002classifying}
M.~Varma and A.~Zisserman.
\newblock Classifying images of materials: Achieving viewpoint and illumination
  independence.
\newblock In {\em European Conference on Computer Vision}, pages 255--271.
  Springer, 2002.

\bibitem{wang2016dataset}
T.-C. Wang, J.-Y. Zhu, E.~Hiroaki, M.~Chandraker, A.~Efros, and R.~Ramamoorthi.
\newblock A {4D} light-field dataset and {CNN} architectures for material
  recognition.
\newblock In {\em Proceedings of European Conference on Computer Vision
  (ECCV)}, 2016.

\bibitem{xiao2010sun}
J.~Xiao, J.~Hays, K.~A. Ehinger, A.~Oliva, and A.~Torralba.
\newblock Sun database: Large-scale scene recognition from abbey to zoo.
\newblock In {\em Computer vision and pattern recognition (CVPR), 2010 IEEE
  conference on}, pages 3485--3492. IEEE, 2010.

\bibitem{xue16}
J.~Xue, H.~Zhang, K.~Dana, and K.~Nishino.
\newblock Differential angular imaging for material recognition.
\newblock In {\em ArXiv}, 2016.

\bibitem{yu2015lsun}
F.~Yu, A.~Seff, Y.~Zhang, S.~Song, T.~Funkhouser, and J.~Xiao.
\newblock Lsun: Construction of a large-scale image dataset using deep learning
  with humans in the loop.
\newblock {\em arXiv preprint arXiv:1506.03365}, 2015.

\bibitem{zhang15}
H.~Zhang, K.~Dana, and K.~Nishino.
\newblock Reflectance hashing for material recognition.
\newblock {\em IEEE Conference on Computer Vision and Pattern Recognition
  (CVPR)}, pages 3071--3080, 2015.

\bibitem{zhang16}
H.~Zhang, K.~Dana, and K.~Nishino.
\newblock Friction from reflectance: Deep reflectance codes for predicting
  physical surface properties from one-shot in-field reflectance.
\newblock In {\em Proceedings of the European Conference on Computer Vision
  (ECCV)}, 2016.

\bibitem{zhao2015stacked}
J.~Zhao, M.~Mathieu, R.~Goroshin, and Y.~Lecun.
\newblock Stacked what-where auto-encoders.
\newblock {\em arXiv preprint arXiv:1506.02351}, 2015.

\end{thebibliography}
}

\end{document}